\documentclass[runningheads]{llncs}
\usepackage[T1]{fontenc}
\usepackage{lipsum}
\usepackage{framed}
\usepackage{color}
\usepackage{amsfonts}
\usepackage{amsmath}
\usepackage{amssymb}
\usepackage{graphicx}     
\usepackage{capt-of}
\usepackage{graphics}
\usepackage{graphicx}
\usepackage{multicol}
\usepackage{multirow}
\usepackage{colortbl} 
\usepackage{url}
\usepackage{hyperref}
\usepackage{siunitx} 
\begin{document}
\title{HAVE-Net: Hallucinated Audio-Visual Embeddings for Few-Shot Classification with Unimodal Cues}
%
%
\author{Ankit Jha\inst{1}\orcidID{0000-0002-1063-8978} \and
Debabrata Pal\inst{1}\orcidID{0000-0003-2667-6642} \and\\ Mainak Singha\inst{1}\and
Naman Agarwal\inst{1}\and Biplab Banerjee\inst{1}\orcidID{0000-0001-8371-8138}}
\authorrunning{A. Jha et al.}
%
\institute{Indian Institue of Technology Bombay,
India
\email{\{ankitjha16,deba.iitbcsre19,mainaksingha.iitb,\\naman.agarwal319,getbiplab\}@gmail.com}}
\maketitle              
\begin{abstract}
Recognition of remote sensing (RS) or aerial images is currently of great interest, and advancements in deep learning algorithms added flavor to it in recent years. Occlusion, intra-class variance, lighting, etc., might arise while training neural networks using unimodal RS visual input. Even though joint training of audio-visual modalities improves classification performance in a low-data regime, it has yet to be thoroughly investigated in the RS domain. Here, we aim to solve a novel problem where both the audio and visual modalities are present during the meta-training of a few-shot learning (FSL) classifier; however, one of the modalities might be missing during the meta-testing stage. This problem formulation is pertinent in the RS domain, given the difficulties in data acquisition or sensor malfunctioning. To mitigate, we propose a novel few-shot generative framework, \textit{Hallucinated Audio-Visual Embeddings-Network (HAVE-Net)}, to meta-train cross-modal features from limited unimodal data. Precisely, these hallucinated features are meta-learned from \textit{base} classes and used for few-shot classification on \textit{novel} classes during the inference phase. The experimental results on the benchmark ADVANCE and AudioSetZSL datasets show that our hallucinated modality augmentation strategy for few-shot classification outperforms the classifier performance trained with the real multimodal information at least by 0.8-2\%.


\end{abstract}

\section{Introduction}
\begin{figure}[ht!]
    \centering
    \includegraphics[width=9cm]{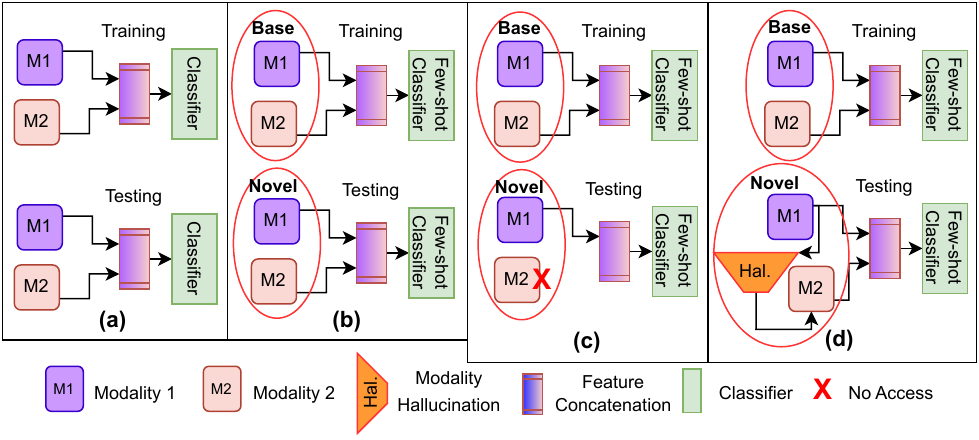}
    \vspace*{-0.5cm}
    \caption{\scriptsize{Illustration of (a) Fully supervised classifier considering same classes during the training and testing phases, (b) Multimodal few-shot classification, and (c) Issues with non-availability of a single modality for the novel classes during testing. To mitigate (d) we propose modality hallucination for multimodal audio-visual few-shot classification. The hallucinated modality generator is trained on \textit{base} (seen) classes and used only at the testing time on the novel (unseen) classes.}}
    \label{fig:teaser}
    \vspace*{-4mm}
\end{figure}

Accurate remote sensing (RS) image classification with fewer training samples is crucial for land-cover classification, voice-to-satellite image retrieval, efficient route planning, etc. While existing researches primarily rely on optical images or unimodal data, incorporating audio cues can tackle several real-world challenges. For example, audio can assist in locating visually obscured objects or identifying tiny features in low-resolution satellite images. However, the lack of paired audio-visual training data hinders the development of reliable multimodal systems. Although there exists limited usage of audio data in RS scene classification, integrating land-cover class-specific audio information with geo-tagged \cite{DBLP:journals/corr/abs-2108-00688} aerial images can significantly boost the RS classification performance.

Metric learning \cite{koch2015siamese,snell2017prototypical} and optimization-based meta-learning \cite{DBLP:conf/nips/FinnXL18} are common approaches in the few-shot learning (FSL) \cite{matching-net,DBLP:conf/nips/FinnXL18,koch2015siamese,snell2017prototypical} community. Siamese networks \cite{koch2015siamese} extract the image pair features using shared parameters and seeks to build an efficient distance metric based on feature similarity. In Prototypical Networks \cite{snell2017prototypical}, the distance from the nearest class prototype is used to determine the label of a test query. On the other hand, MAML \cite{DBLP:conf/nips/FinnXL18} focuses on learning initial parameters for rapid adaptation to unobserved tasks. Besides, SPN \cite{9455864} employs Bayesian approximation to stabilize model uncertainty in FSL. 

Due to the poor perception of the classification environment by uni-modal learning, researchers \cite{pahde2021multimodal} further investigated multimodal learning in a few-shot context. For example, Multimodal Prototypical Networks \cite{pahde2021multimodal} additionally considers hallucinated text modality. In literature \cite{hu2020crosstask,9093438}, Multimodal learning has demonstrated significant performance gain. However, accessing all the modalities during testing is difficult and can impact classification performance. We address this problem by hallucinating all the unavailable modalities during test time. 


In this paper, we propose to use a Conditional Generative Adversarial Network (CGAN) \cite{cgan} to hallucinate missing modality data. Unlike traditional CGAN, we train a modality and class-conditioned Conditional Multimodal GAN (CMM-GAN) by episodic meta-learning with \textit{base} class samples, allowing it to generate cross-modal features for novel classes. During meta-training, we utilize both audio and visual modality data to extract discriminative features with strong cross-modal correlation (Figure \ref{fig:teaser}). The hallucinated features are augmented with available real modality features to enrich the distribution density. Then, during meta-testing, our approach involves hallucinating missing modality data  based on the available real modality data of \textit{base} (seen) classes. These hallucinated features are utilized to perform few-shot multimodal classification on novel (unseen) classes. Our main contributions are as follows:

\begin{figure*}[http!]
  \begin{center}
  \includegraphics[width=\linewidth]{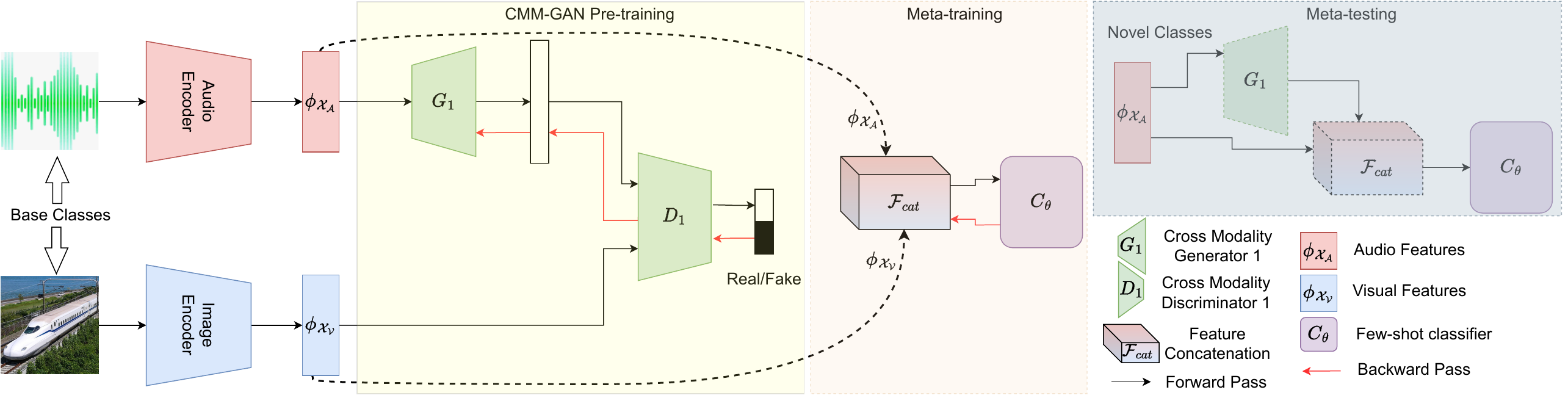}
  \vspace*{-8mm}
  \caption{\scriptsize{The overview of our proposed HAVE-Net framework. We first pre-train CMM-GAN, composed of a Generator $G_1$ and a Discriminator $D_1$ for audio to visual cross-modal feature generation using a large number of \textit{base} class samples. Then, we meta-train a few-shot learning classifier $C_{\theta}$ using the multimodal dataset (original fused modality on \textit{base} class limited samples). During Meta-testing, the classification is performed for the novel class samples based on the available modality features augmented with the hallucinated cross-modal features from the pre-trained CMM-GAN.}}
    \label{fig:block_dig}
  \end{center}
  \vspace*{-4mm}
\end{figure*}


\noindent- We introduce a novel modality-agnostic multimodal framework in a few-shot regime, namely, HAVE-Net, to battle against the missing modality problem in the meta-testing phase.\\
- We propose a novel Conditional Multimodal Adversarial loss objective to pre-train our CMM-GAN classifier from the available unimodal data (either audio to visual or vice-versa) \\
- We conduct extensive experiments on the benchmark AudioSetZSL \cite{9093438} and ADVANCE \cite{hu2020crosstask} datasets, showcasing our method's superior few-shot multimodal classification performance from available unimodal data.

\section{Related Works}
\vspace{-0.3cm}
\noindent{\textbf{A. Multimodal Learning:}}  
RS audiovisual deep learning has received limited attention in existing research. However, some works have been proposed in this area like, \cite{8486338} focuses on learning the correspondence between the audio and visual modalities for cross-modal retrieval of RS images. Similarly, \cite{8517977} proposes a clustering-based \textit{aural atlas} approach that fuses audiovisual information. 
Recently, a self-supervised learning-based approach was proposed in \cite{DBLP:journals/corr/abs-2108-00688} to understand the key mapping between RS audiovisual samples, which was extended to other transfer learning tasks such as scene classification \cite{hu2020crosstask}, semantic segmentation, and cross-modal retrieval. In \cite{hu2020crosstask}, sound-image pairs were enforced to transfer sound event information for RS scene classification. It is worth noting that the existing models are designed for specific pairs of modalities. 
\\
\noindent{\textbf{B. Few-shot Learning:}} To overcome the challenge of limited labeled data in remote sensing (RS) classification, researchers have proposed various methods, such as semi-supervised CNN, self-taught learning, and few-shot learning (FSL) \cite{sung2018learning,matching-net}. FSL uses the ``learning to learn" approach, where the model is trained on fully-supervised \textit{base} classes to transfer learning to novel, label-deficient categories \cite{DBLP:conf/nips/FinnXL18,ren2018meta}. The use of meta-learning algorithms, like MAML \cite{DBLP:conf/nips/FinnXL18}, enables the model to learn invariant features by creating few-shot tasks (episodes) for training. Additionally, metric-learning methods \cite{matching-net} aim to learn an optimal feature space with minimal intra-class and maximum inter-class distances.

\section{Methodology}
\vspace{-0.3cm}
\subsection{Preliminaries}
\vspace{-0.3cm}
Let $\mathcal{D = \{X_A, X_V; Y\}}$ be the multimodal dataset, where $\mathcal{X_A}$ is an audio spectrogram, $\mathcal{X_V}$ represents visual space, and $\mathcal{Y}$ is their corresponding label space. Further, let $x{^i}{_A} \in \mathcal{X_A}$ and $x{^i}{_V} \in \mathcal{X_V}$ be the $i^{th}$ input sample point and $y^{i}$ is its associated label.
In FSL, two disjoint sets of classes namely, base classes $\mathcal{Y}_{Base}$ for training and novel classes $\mathcal{Y}_{Novel}$ for testing are sampled from each dataset where, $\mathcal{Y} = \mathcal{Y}_{Base} \cup \mathcal{Y}_{Novel}$ and $\mathcal{Y}_{Base} \cap \mathcal{Y}_{Novel} = \phi$. $\mathcal{Y}_{Base}$ has ample amount of training sample points such that we assert $|\mathcal{Y}_{Base}| > |\mathcal{Y}_{Novel}|$. As the few-shot classifier trained using original modalities, $\mathcal{X_A}$ and $\mathcal{X_V}$, the presence of all the modalities certainly boosts the FSL classifier performance. However, we assume only one modality during testing, causing the classifier to underperform for the novel classes. So to make the missing modality available at the inference, we pre-train CMM-GAN over all samples of the base classes to generate the cross-modal features, i.e., either $\mathcal{X_A}$ to $\mathcal{X_V}$ or vice-versa.
During meta-training, random episodes are sampled consisting of support $\mathcal{S}$ and query set $\mathcal{Q}$. The support set is defines as $\mathcal{S} = \{(x{^i}{_A}, x{^i}{_V}; y^{i})\}_{i=1}^{m\mathcal{K}}$ where $m$ training samples per $\mathcal{K} \in {\mathcal{Y}}_{Base}$ classes are sampled in each episode and also known as $\mathcal{K}$-way $m$-shot classification. The query set $\mathcal{Q} = \{(x{^j}{_A}, x{^j}{_V}; y^{j})\}_{j=1}^{n\mathcal{K}}$ is formed with other disjoint random $n$ samples per $\mathcal{K}$ class. By training from a wide range of $\mathcal{Y}_{Base}$ samples, classifier $C_{\theta}$ gets transferrable knowledge to classify single modality data. Also, CMM-GAN learns to generate pseudo samples of cross-modal data. In this paper, during meta-training, we use both $\mathcal{X_A, X_V}$ whereas, during meta-testing, we assume one of the modalities is absent. Hence, we generate pseudo samples from other available modalities using CMM-GAN and classify samples belong to the $\mathcal{Y}_{Novel}$.

\vspace*{-4mm}
\subsection{Proposed Methodology}
\vspace{-2mm}
In Figure \ref{fig:block_dig}, we show the architectural diagram of our proposed HAVE-Net. Here, we explain the two phases adopted in training the HAVE-Net, i.e., i) base training for cross-modal hallucination, which is the main essence of our proposed setup, and ii) incrementally training the multimodal few-shot classifier. To begin with, we first use the VGGish model pre-trained on Audio Set \cite{45857} dataset as audio encoder to extract the audio features $\phi_{\mathcal{X_A}} \in {\mathbb{R}}^{1\times 1024}$ for each of the audio spectrograms $x{^i}{_A}$. Similarly, for each $x{^i}{_V}$ we extract visual features $\phi_{\mathcal{X_V}} \in {\mathbb{R}}^{1\times 1024}$ using pre-trained ResNet model on remote sensing RESISC45 \cite{RESISC45} dataset. Note that we haven't used pre-trained ImageNet features due to wide domain differences with RS scenes.\\
\noindent{\textbf{Base Training:}} We split this phase into two parts; firstly, CMM-GAN is trained for cross-modal feature hallucination, where generator $G_1$ of CMM-GAN generates base class visual features $\phi_{\mathcal{X_V}}$ from audio features $\phi_{\mathcal{X_A}}$. Similarly, using $G_2$, we generate audio features from visual features. Inspired by \cite{pahde2021multimodal}, we train CMM-GAN by imposing the constraints over one modality to generate other modality pseudo features. Mathematically, 
CMM-GAN is pre-trained on the large number of base class audio spectrogram features $\tiny{\underset{Base}{G_1}(\phi_{\mathcal{X_A}}) \approx \underset{Base}{\phi_{\mathcal{X_V}}}}$ and thereafter it is used to generate the pseudo-visual features from limited novel audio class samples $\underset{Novel}{\phi_{\mathcal{X_V}}}$ $=$ ${\underset{Base}{G_1}(\phi_{{\underset{Novel}{\mathcal{X_{A}}}}})}$ in testing or vice-versa. Secondly, we perform training of FSL classifier using multimodal real audio-visual features. We fuse the audio and visual features ($\mathcal{X_A}$ and $\mathcal{X_V}$) as: $\mathcal{F}_{cat}$ $=$ $[\phi_{\mathcal{X_A}} ; \phi_{\mathcal{X_V}}]$, where `;' indicates the concatenation operator. We train $\mathcal{F}_{cat}$ in a similar fashion as in SPN \cite{9455864} to stabilize the model predictions. The classifier ${C_{\theta}}$ aims to learn from a small set of audio and visual features from limited base class samples ${C_{\theta}(\phi_{\mathcal{F}_{cat}})}$, where $C_{\theta}$ denotes the FSL classifier with its learnable parameters $\theta$.

\noindent{\textbf{Novel Testing:}} 
We concatenate pre-trained CMM-GAN hallucinated visual features conditioned on only available novel unimodal real audio features to mitigate the need for missing visual data. Then multimodal few-shot classification is performed by the meta-trained classifier, $C_{\theta}(\phi_{[\mathcal{X_A} ; G_1(\mathcal{X_A})]})$. Similarly, from novel visual samples and corresponding hallucinated audio samples, we perform multimodal classification $C_{\theta}(\phi_{[\mathcal{X_V} ; G_2(\mathcal{X_V})]})$. This helps in shifting the initial uncertain class prototype to the true distribution centroid position in the metric space.
\vspace*{-3mm}
\subsection{{Objective Functions}}
\vspace*{-2mm}
\label{sec:object}
Here, we define the three loss components in optimizing the proposed HAVE-Net, which are as follows:\\
\noindent{\textbf{CMM-GAN Loss:}} We optimize CMM-GAN using conditional multimodal adversarial loss $\mathcal{L}_{CMA}$ comprising equations \ref{gan_loss} and \ref{reco} to hallucinate visual features from the available audio features.
\vspace*{-2mm}
\begin{equation}
    \begin{array}{cc}
        \mathcal{L}_{CGAN}(G_1,D_1) = \underset{G_1}{min} \hspace{1mm} \underset{D}{max}\hspace{1mm} \mathbb{E}_{y \sim \phi_{\mathcal{X}_V}} [log D_1(y)]&  \\
      + \mathbb{E}_{z \sim P_z, \mathcal{X}_{A}}[log (1-D_1) (G_1(\phi_{\mathcal{X}_{A}}|\phi_{\mathcal{X}_{V}},z))]   & 
    \end{array}
    \label{gan_loss}
    \vspace*{-2.2mm}
\end{equation}
Additionally, we introduce reconstruction loss $\mathcal{L}_{Rec}$ in (Eq. \ref{reco}) in defining $\mathcal{L}_{CMA}$, which helps in minimizing the distance between the generated visual features and corresponding ground truth. The overall CMM-GAN objective function is defined in (Eq.\ref{OO}) where $\lambda_1$ is a weight factor.
\vspace*{-2.2mm}
\begin{gather}
\mathcal{L}_{Rec}(G_1)= ||\phi_{\mathcal{X}_{V}}-G_{1}(\phi_{\mathcal{X}_{A}})|| \label{reco}\\
\mathcal{L_{CMA}}(D_1)= \mathcal{L}_{CGAN}(G_1,D_1) + \lambda_{1} \cdot \mathcal{L}_{Rec}(G_1)
\label{OO}
\end{gather}
Similarly, we hallucinate visual features from the available audio features using generator $G_2$ and discriminator $D_2$.   

\noindent{\textbf{Prototypical Loss:}} In each episode, $(\mathcal{S, Q}) \subset \mathcal{D}$ is randomly sampled under FSL setting. We calculate the training samples' multimodal (audio-visual) prototypes $c_k$ for $k$ true classes, which is the mean of the support audio-visual features $(x: [\phi_{\mathcal{X}_{A}}, \phi_{\mathcal{X}_{V}}])$ and later used to find the nearest prototype for the test samples based on the euclidean distance function.
\vspace*{-2.2mm}
\begin{equation}
  \mathcal{L}_{Proto}= \dfrac{\exp(-d(C_\phi (x),c_k)}{{\sum_{k^{'}=1}^{\mathcal{K}}} \exp(-d(C_\phi (x),c_{k^{'}})}
\end{equation}

\noindent{\textbf{FSL uncertainty minimization loss:}} For $i^{th}$ query image, standard deviation and mean across all prediction probabilities $p_{i,multiple} = \{p_i^1,p_i^2,p_i^3,\cdots,p_i^{n_{times}}\}$ are $\sigma_{Loss,i}$, and, $\mu_p = \frac{\sum_{y=1}^{n_{times}} p_{ik}^y}{n_{times}}$. The prediction variance loss for all queries, 
\begin{equation}\label{StdLoss}
\begin{array}{cc}
  \mathcal{L}_{std} = \sum_{i=1}^{n\mathcal{K}} \sigma_{Loss,i} = \sum_{i=1}^{n\mathcal{K}} \sqrt{\frac{\sum_{y=1}^{n_{times}}(p_{ik}^y - \mu_p)^2}{n_{times}}}&\\
\end{array}
\end{equation}
\noindent{\textbf{Total Loss:}} In Eq. 6, to train the HAVE-Net, we jointly optimize the CMM-GAN (Eq. 3) and the prototypical loss (Eq. 4) with the uncertainty minimization loss $\mathcal{L}_\text{std}$ (Eq. 5), where $\lambda_2$ represents the hyperparameter.
\begin{equation} 
\mathcal {L}_{\text {Total}} = \mathcal{L_{CMA}} + \mathcal {L}_{Proto} + \lambda_2 \times \mathcal {L}_{\text {std}}
\end{equation} 

\vspace{-8mm}
\section{Experiments and Analysis}
\vspace*{-2mm}
\subsection{Datasets} 
\vspace*{-2mm}
We evaluated the proposed method on two benchmark multimodal (audio-visual) datasets namely, AudioSetZSL \cite{9093438} \& ADVANCE \cite{hu2020crosstask} datasets. The AudioSetZSL is a subset of Audio Set \cite{45857} contains weakly annotated visuals related to different audio events. We set $|\mathcal{Y}_{Base}| = 23$ and $|\mathcal{Y}_{Novel}| = 10$ as per train-test split in \cite{9093438}, where $|\mathcal{Y}|$ denotes the cardinality of $\mathcal{Y}$. We used the features provided in \cite{9093438} for our few-shot methodologies. On a similar note, the ADVANCE dataset, which has pairs of RS audio and visual modalities and are randomly split into $|\mathcal{Y}_{Base}| = 8$ and $|\mathcal{Y}_{Novel}| = 5$. 
\vspace{-5.5mm}
\begin{table*}[htbp]
\caption{\scriptsize{ Comparison of different meta-learning methods for $m$ ways $\in$ \{3, 5\} with k-shot $\in$ \{1, 5\} on ADVANCE dataset. A + (Hal. V)$^{\textbf{*}}$ is the fusion of audio modality with the hallucinated visual modality. V + (Hal. A)$^{\textbf{**}}$ represents the fusion of visual modality with hallucinated audio modality during the meta-testing.}}
\vspace*{-6mm}
\begin{center}
\scalebox{0.62}{
\begin{tabular}{l|c|c c|c c|c c}
\hline
\textbf{Model}&\textbf{m-way}&\multicolumn{2}{c|}{\textbf{Audio (A)}} &\multicolumn{2}{c}{\textbf{Visual (V)}}&\multicolumn{2}{|c}{\textbf{Fusion}} \\

&&\multicolumn{2}{c}{}&\multicolumn{2}{|c|}{} &\multicolumn{2}{|c}{} \\

\cline{3-8} 
& &\textbf{1-shot}&\textbf{5-shot}&\textbf{1-shot}&\textbf{5-shot}&\textbf{1-shot}&\textbf{5-shot}\\
\hline

Prototypical \cite{snell2017prototypical} &3-way   &56.12$\pm$0.38 &73.35$\pm$0.44     &58.65$\pm$0.24 &77.13$\pm$0.37     &65.44$\pm$0.32 &81.96$\pm$0.34  \\
Relation Net \cite{sung2018learning} &3-way   &39.00$\pm$0.43 &40.45$\pm$0.44     &43.91$\pm$0.22 &47.96$\pm$0.18     &49.08$\pm$0.55 &56.33$\pm$0.37 \\

MAML \cite{DBLP:conf/nips/FinnXL18} &3-way   &57.20$\pm$0.30 &74.61$\pm$0.31     &64.31$\pm$0.25 &79.27$\pm$0.24     &76.60$\pm$0.42 &83.01$\pm$0.31  \\

Matching-Net \cite{matching-net} &3-way   &56.50$\pm$0.48 &69.53$\pm$0.36     &60.09$\pm$0.32 &76.35$\pm$0.29     &70.11$\pm$0.39 &81.60$\pm$0.41    \\


HAVE-Net(\textbf{A} + Hal. $\textbf{V})^\textbf{*}$  & 3-way&59.41$\pm$0.55 &75.03$\pm$0.12     &67.42$\pm$0.35 &81.50$\pm$0.31&\textbf{\color{blue} 79.61$\pm$0.34}&\textbf{\color{blue}84.75$\pm$0.19} \\
HAVE-Net(\textbf{V} + Hal. $\textbf{A})^\textbf{**}$  & 3-way&59.41$\pm$0.55 &75.03$\pm$0.12     &67.42$\pm$0.35 &81.50$\pm$0.31& \textbf{\color{blue}77.43$\pm$0.27} &\textbf{\color{blue}84.30$\pm$0.45} \\
\hline 

Prototypical \cite{snell2017prototypical} &5-way   &43.63$\pm$0.32 &68.20$\pm$0.26     &46.32$\pm$0.57 &70.47$\pm$0.33     &55.71$\pm$0.29 &73.69$\pm$0.48\\
Relation Net \cite{sung2018learning} &5-way   &28.72$\pm$0.38 &31.93$\pm$0.46     &33.66$\pm$0.22 &36.80$\pm$0.63     &39.61$\pm$0.32 &44.36$\pm$0.41  \\

MAML \cite{DBLP:conf/nips/FinnXL18} &5-way   &47.31$\pm$0.19 &71.52$\pm$0.33     &48.97$\pm$0.42 &72.65$\pm$0.19     &59.00$\pm$0.56 &75.82$\pm$0.34 \\

Matching-Net \cite{matching-net} &5-way   &43.87$\pm$0.40 &69.32$\pm$0.24     &47.15$\pm$0.30 &66.90$\pm$0.39     &56.41$\pm$0.50& 71.20$\pm$0.30    \\


HAVE-Net(\textbf{A} + Hal. $\textbf{V})^\textbf{*}$ & 5-way&53.70$\pm$0.31 &74.29$\pm$0.24     &56.22$\pm$0.35 &76.13$\pm$0.46& \textbf{\color{blue}61.39$\pm$0.48}&\textbf{\color{blue}76.80$\pm$0.26 } \\
HAVE-Net(\textbf{V} + Hal. $\textbf{A})^\textbf{**}$  & 5-way&53.70$\pm$0.31 &74.29$\pm$0.24     &56.22$\pm$0.35 &76.13$\pm$0.46& \textbf{\color{blue}60.54$\pm$0.18}& \textbf{\color{blue}76.32$\pm$0.40}\\
\hline
\multicolumn{8}{l}{We highlight the best results in \textbf{\color{blue}{blue}}, where it represents HAVE-Net's performance during inference time.}\\ 
\end{tabular}}
\label{tab_1shot}
\end{center}
\vspace{-8mm}
\end{table*}

\vspace{-4mm}
\subsection{Training Protocol}
\vspace{-2mm}
As described in Section 2.2, we extract the audio and visual features ($\mathbb{R}^{1\times 1024}$) from the pre-defined encoders and train the CMM-GAN model for cross-modal feature hallucination and FSL classifiers for the classification task.\\
\noindent \textbf{CMM-GAN Pre-training:} For AudioSetZSL, the audio features are readily available in \cite{9093438} whereas for ADVANCE dataset, we extract the audio features from ResNet18 \cite{DBLP:journals/corr/HeZRS15} trained on RESICS45 \cite{RESISC45} dataset.
We pre-train CMM-GAN with a batch size of 128 and set Adam optimizer with a learning rate of $10^{-4}$ with $\lambda_1$ to $1$ in (Eq. \ref{OO}).\\
\noindent{\textbf{Few-shot Multimodal Training:}} We randomly sample base class data at $m$ $\in$ $\{5$, $10\}$ and $m$ $\in$ $\{3$, $5\}$ ways respectively for AudioSetZSL \cite{9093438} and ADVANCE \cite{hu2020crosstask} datasets with $\mathcal{K}$ $\in$ $\{1$, $5\}$. We choose $\lambda_2$ (Eq. 5) to be 10 and optimize few-shot classifier of HAVE-Net using Adam optimizer with learning rate $10^{-3}$. We train the model for 60 epochs with 100 episodes per epoch.
\vspace{-4mm}
\subsection{Discussion}
\vspace{-2mm}
\noindent{\textbf{Competitors:}} 
To validate the performance of our proposed framework, we compare the results for the ADVANCE and AudioSetZSL datasets using different state-of-the-art (SOTA) meta-learning methods in Tables \ref{tab_1shot} and \ref{tab_1shot_adv}, respectively \footnote{To the best of our knowledge, we compare our proposed method with most relevant method. \cite{pahde2021multimodal} concentrates on image-text pairs as multiple modalities, which is not fair to use in comparison with our problem of joint audio-visual learning.} The dominating modalities in both datasets are distinct, i.e., audio in AudioSetZSL and visual in ADVANCE; the FSL classifiers can indeed classify their joint representation better than separate ones.
\vspace*{-5mm}
\begin{table*}[htbp]
\caption{\scriptsize{ Comparison of different meta-learning methods for $m$ ways $\in$ \{5, 10\} with k-shot $\in$ \{1, 5\} on AudioSetZSL dataset. A + (Hal. V)$^{\textbf{*}}$ is the fusion of audio modality with the hallucinated visual modality. V + (Hal. A)$^{\textbf{**}}$ represents the fusion of visual modality with hallucinated audio modality during the meta-testing. }}
\vspace*{-6mm}
\begin{center}
\scalebox{0.62}{
\begin{tabular}{l|c|c c|c c|c c}
\hline
\textbf{Model}&\textbf{m-way}&\multicolumn{2}{c|}{\textbf{Audio (A)}} &\multicolumn{2}{c}{\textbf{Visual (V)}}&\multicolumn{2}{|c}{\textbf{Fusion}}  \\

&&\multicolumn{2}{c}{}&\multicolumn{2}{|c|}{} &\multicolumn{2}{|c}{}  \\
\cline{3-8} 
& &\textbf{1-shot}&\textbf{5-shot}&\textbf{1-shot}&\textbf{5-shot}&\textbf{1-shot}&\textbf{5-shot}
\\
\hline

Prototypical \cite{snell2017prototypical} &5-way   &50.56$\pm$0.31 &67.18$\pm$0.24     &43.39$\pm$0.35 &62.26$\pm$0.31     &61.30$\pm$0.53 &76.11$\pm$0.43\\
Relation Net \cite{sung2018learning} &5-way   &50.47$\pm$0.34 &55.69$\pm$0.41     &42.83$\pm$0.27 &50.45$\pm$0.32     &60.75$\pm$0.35 &66.75$\pm$0.33\\

MAML \cite{DBLP:conf/nips/FinnXL18} &5-way   &51.89$\pm$0.21 &66.93$\pm$0.23     &44.57$\pm$0.22 &63.30$\pm$0.21    &60.80$\pm$0.21 &77.65$\pm$0.31 \\
Matching-Net \cite{matching-net} &5-way   &50.60$\pm$0.36 &62.76$\pm$0.40     &42.91$\pm$0.30 &63.08$\pm$0.43     &59.40$\pm$0.51 &70.22$\pm$0.48   \\ 

HAVE-Net(\textbf{A} + Hal. $\textbf{V})^\textbf{*}$  & 5-way&53.70$\pm$0.44 &69.95$\pm$0.32     &46.14$\pm$0.16 &65.97$\pm$0.31&\textbf{\color{blue}62.38$\pm$0.30}&\textbf{\color{blue}78.86$\pm$0.29} \\
HAVE-Net(\textbf{V} + Hal. $\textbf{V})^\textbf{**}$  & 5-way&53.70$\pm$0.44 &69.95$\pm$0.32     &46.14$\pm$0.16 &65.97$\pm$0.31&\textbf{\color{blue}61.89$\pm$0.22 }&\textbf{\color{blue}78.31$\pm$0.54} \\
\hline 

Prototypical \cite{snell2017prototypical} &10-way   &34.09$\pm$0.27 &51.66$\pm$0.45     &28.39$\pm$0.47 &45.23$\pm$0.36    &43.13$\pm$0.31 &60.29$\pm$0.41     \\
Relation Net \cite{sung2018learning} &10-way   &31.39$\pm$0.27 &37.86$\pm$0.34     &26.30$\pm$0.32 &30.67$\pm$0.28     &39.39$\pm$0.54 &42.17$\pm$0.41     \\

MAML \cite{DBLP:conf/nips/FinnXL18} &10-way   &34.35$\pm$0.63 &48.91$\pm$0.43     &32.99$\pm$0.36 &47.68$\pm$0.32     &44.19$\pm$0.24 &63.87$\pm$0.18    \\
Matching-Net \cite{matching-net} &10-way   &32.15$\pm$0.43 &46.40$\pm$0.31     &29.73$\pm$0.25 &42.60$\pm$0.24     &42.57$\pm$0.42 &61.22$\pm$0.31    \\


HAVE-Net(\textbf{A} + Hal. $\textbf{V})^\textbf{*}$ & 10-way&36.69$\pm$0.35 &53.02$\pm$0.15     &32.11$\pm$0.31 &48.91$\pm$0.22& \textbf{\color{blue}49.07$\pm$0.39}&\textbf{\color{blue}66.00$\pm$0.27 }\\
HAVE-Net(\textbf{V} + Hal. $\textbf{A})^\textbf{**}$ & 10-way&36.69$\pm$0.35 &53.02$\pm$0.15     &32.11$\pm$0.31 &48.91$\pm$0.22& \textbf{\color{blue}48.75$\pm$0.14}&\textbf{\color{blue}65.18$\pm$0.33} \\
\hline
\multicolumn{8}{l}{We highlight the best results in \textbf{\color{blue}{blue}}, where it represents HAVE-Net's performance during inference time.}\\ 
\end{tabular}}
\label{tab_1shot_adv}
\end{center}
\vspace{-10mm}
\end{table*}

On the ADVANCE dataset, HAVE-Net with audio and hallucinated visual features (Audio + Hal. Visual) and (Visual + Hal. Audio) significantly outperforms the other SOTA approaches for the fusion of original modalities at least by 1.1 \% on 3-way and 0.7 \% on 5-way classification settings. While on the AudioSetZSL dataset, our HAVE-Net outperforms the referred methods on 5-way and 10-way classification settings by at least 0.8 \% and 2 \%, respectively. 
We show the ablation results in supplementary paper.

\vspace{-6mm}
\section{Takeaways}
\vspace{-4.5mm}
We address the challenge of remote sensing audio-visual FSL in the absence of unimodal data during testing. To overcome this limitation, we utilized CMM-GAN to generate cross-modality features from the available unimodal data of base classes, which were then used to hallucinate the missing modality for novel classes during inference. Our proposed modal-agnostic unified framework, HAVE-Net, is designed to overcome the limitations of existing few-shot learning methods and is particularly useful in scenarios where robotic sensors may malfunction or data acquisition may be limited. We evaluated the performance of HAVE-Net against several state-of-the-art meta-learning algorithms on benchmark datasets, i.e., ADVANCE and AudioSetZSL, and our results demonstrate its favorable performance compared to using actual modalities during testing.

%
%
%
\vspace{-4mm}
\bibliographystyle{splncs04}
\bibliography{egbib}
%




\end{document}


\title{Supplementary for HAVE-Net: Hallucinated Audio-Visual Embeddings for Few-Shot Classification with Unimodal Cues}
%
%
\author{Ankit Jha\inst{1}\orcidID{0000-0002-1063-8978} \and
Debabrata Pal\inst{1}\orcidID{0000-0003-2667-6642} \and Mainak Singha\inst{1}\and
Naman Agarwal\inst{1}\and Biplab Banerjee\inst{1}\orcidID{0000-0001-8371-8138}}
\authorrunning{A. Jha et al.}
%
\institute{Indian Institue of Technology Bombay,
India
\email{\{ankitjha16,deba.iitbcsre19,mainaksingha.iitb,\\naman.agarwal319,getbiplab\}@gmail.com}}
\maketitle              
\section{Additional Information}
\noindent{\textbf{Inference Strategy:}} Once CMM-GAN learns to generate hallucinated visual data and $C_{\theta}$ classifies base class $\mathcal{Q}$, we perform meta-testing with novel samples from only audio queries. Here, hallucinated visual queries are generated first and followed by multimodal classification $C_{\theta}(\phi_{[\mathcal{X_A} ; G_1(\mathcal{X_A})]})$. A similar objective function is applicable when we perform multimodal classification from available visual queries.
\section{Ablation Analysis}
\begin{figure*}[!htbp]
    \centering
    \includegraphics[height=3cm]{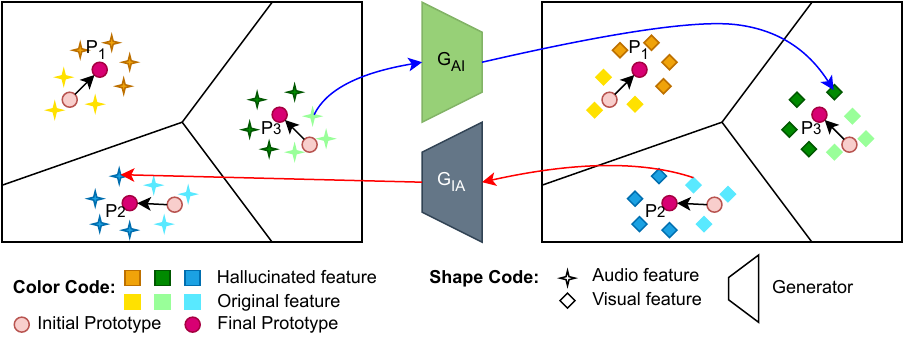}  
    \vspace*{-5mm}
    \caption{{Generator ${G}_{1}$ converts original audio features to hallucinate visual features. Similarly, ${G}_{2}$ synthesizes hallucinated audio from visual features. The hallucinated features mitigate data scarcity and shift the initial prototype to a true class centroid position in metric space, boosting the accuracy.}}
    \label{fig:Illustration1}
    \vspace*{-6.5mm}
\end{figure*}

In our proposed HAVE-Net, we concatenate pre-trained CMM-GAN hallucinated visual features conditioned on only available novel unimodal real audio features to mitigate the need for missing visual modality. Then multimodal few-shot classification is performed by the meta-trained classifier, $C_{\theta}(\phi_{[\mathcal{X_A} ; G_1(\mathcal{X_A})]})$. Similarly, from novel visual samples and corresponding hallucinated audio samples, we perform multimodal classification $C_{\theta}(\phi_{[\mathcal{X_V} ; G_2(\mathcal{X_V})]})$. This helps in shifting the initial uncertain class prototype to the true distribution centroid position in the metric space, which is shown in Figure \ref{fig:Illustration1}.


\subsection{Discussion}
\begin{figure*}[htbp!]
  \begin{center}
  \includegraphics[height=4cm]{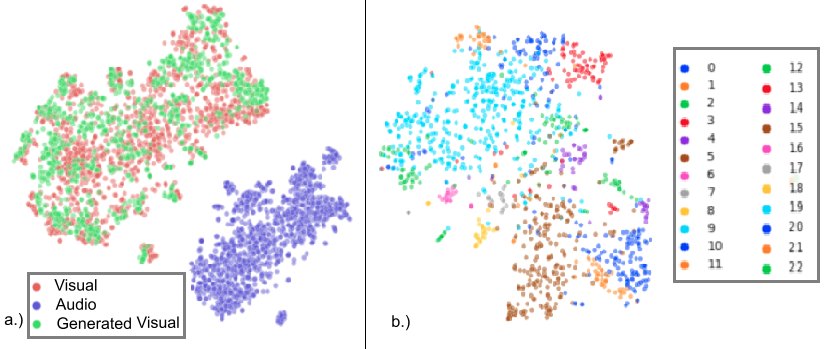} 
  \caption{{The t-SNE plot of a.) (audio, visual, and hallucinated visual) modalities shows that the hallucinated visual samples captured ground truth visual distribution, b.) The 23 hallucinated base class distribution of the visual samples of the AudioSetZSL dataset.}}
  \label{fig:tsne}
  \end{center}
\end{figure*}

\noindent{\textbf{Visualization:}} We use t-SNE \cite{DBLP:journals/corr/abs-1905-02417} to plot the features between the original audio and visual modalities, and cross-modal generated visual modality, shown in Figure \ref{fig:tsne} (a). We observe that the generated visual features from the audio features almost capture the distribution of original visual features. We also presented the t-SNE plot for the class-wise generated visual features ($G_{1}(\phi_{\mathcal{X_A}})\xrightarrow{}\phi_{\mathcal{X_V}}$) using CMM-GAN in Figure \ref{fig:tsne} (b).

\begin{figure}
    \centering
    \includegraphics[width=0.49\textwidth]{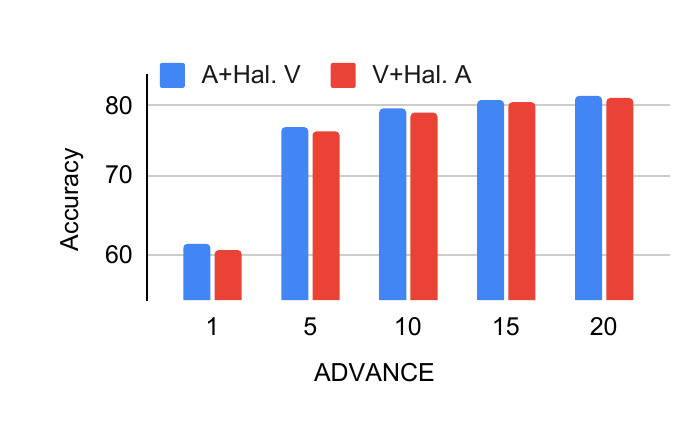}
    \includegraphics[width=0.5\textwidth]{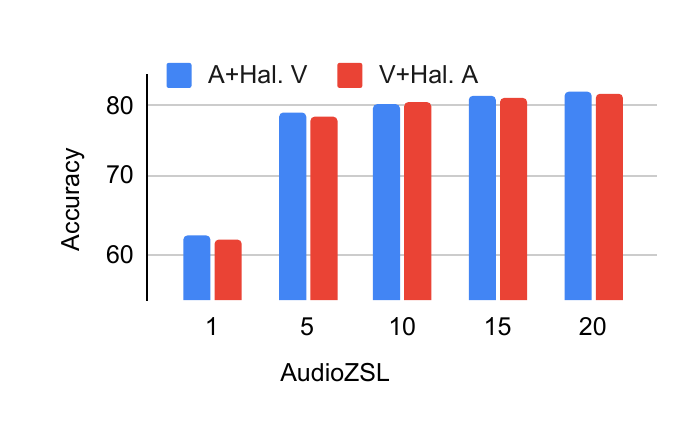}
    
    \caption{Ablation on varying number of shots with 5-way setting for HAVE-Net on ADVANCE and AudioZSL datasets. A + Hal. V is the fusion of audio modality with the hallucinated visual modality. V + Hal. A represents the fusion of visual modality with hallucinated audio modality during the meta-testing.}
    \label{fig:abl_shots}
\end{figure}

\noindent{\textbf{Ablation on Number of Shots:}} To evaluate the effectiveness of HAVE-Net on different amounts of data, we performed an ablation study with varying numbers of shots (i.e., 1, 5, 10, 15, and 20) on 5-way setting. The resulting bar plots are presented in Figure \ref{fig:abl_shots}. Our findings indicate a significant improvement in the classification performance of HAVE-Net for both the fusion of audio modality with the hallucinated visual modality (A + Hal. V) and the fusion of visual modality with the hallucinated audio modality (V + Hal. A) on the ADVANCE and AudioZSL datasets as the number of shots increased.

\begin{table}[!htbp]
\caption{{Ablation study on ADVANCE dataset using different meta-learning methods for $m$ ways $\in$ \{3, 5\} with k-shot $\in$ \{1, 5\}. $\mathcal{X_V}$ and $G_1$($\mathcal{X_A}$) represent the original visual and CMM-GAN hallucinated visual modalities, respectively.}}
\label{abl_adva}
\begin{center}
\scalebox{0.63}{
\begin{tabular}{l|c|c c|c c}
\hline
\textbf{Model}&\textbf{m-way}&\multicolumn{2}{c}{Train on $\mathcal{X_V}$ : Test on ${G_1}$$(\mathcal{X_A})$} 
&\multicolumn{2}{|c}{Train on $G_1$$(\mathcal{X_A})$ : Test on $G_1$$(\mathcal{X_A})$}\\
\cline{3-6} 
& &\textbf{1-shot}&\textbf{5-shot}&\textbf{1-shot}&\textbf{5-shot}\\
\hline
\hline

Prototypical \cite{snell2017prototypical} &3-way    &51.30$\pm$0.57 &76.11$\pm$0.46     &50.73$\pm$0.31 &68.85$\pm$0.22 \\

Relation Net \cite{sung2018learning} &3-way   &41.77$\pm$0.29 &46.23$\pm$0.52     &40.15$\pm$0.32 &47.60$\pm$0.41\\

MAML \cite{DBLP:conf/nips/FinnXL18} &3-way    &51.83$\pm$0.27 &63.08$\pm$0.13     &51.55$\pm$0.43 &61.90$\pm$0.49\\
Matching-Net \cite{matching-net} &3-way    &50.46$\pm$0.35 &61.50$\pm$0.29     &49.11$\pm$0.25 &64.79$\pm$0.32\\

HAVE-Net&3-way        &\textbf{\color{blue}{61.13$\pm$0.20}} &\textbf{\color{blue}{77.23$\pm$0.16}} &\textbf{\color{blue}{58.71$\pm$0.22}} &\textbf{\color{blue}{73.01$\pm$0.24}} \\
\hline
Prototypical \cite{snell2017prototypical} &5-way        &44.87$\pm$0.65 &68.05$\pm$0.33 &41.26$\pm$0.28 &61.37$\pm$0.46 \\
Relation Net \cite{sung2018learning} &5-way     &29.01$\pm$0.21 &31.59$\pm$0.37     &26.42$\pm$0.27 &29.67$\pm$0.51\\

MAML \cite{DBLP:conf/nips/FinnXL18} &5-way    &47.54$\pm$0.19 &64.30$\pm$0.42   &45.50$\pm$0.29 &59.73$\pm$0.64\\
Matching-Net \cite{matching-net} &5-way    &45.70$\pm$0.33 &62.39$\pm$0.21     &40.86$\pm$0.33 &58.21$\pm$0.40\\

HAVE-Net&5-way        &\textbf{\color{blue}{57.45$\pm$0.30}} &\textbf{\color{blue}{76.84$\pm$0.23}} &\textbf{\color{blue}{56.20$\pm$0.12}} &\textbf{\color{blue}{73.55$\pm$0.26}} \\
\hline
\multicolumn{4}{l}{We highlight the best results in \textbf{\color{blue}{blue}}.} \\
\end{tabular}}
\end{center}
\end{table}

\begin{table}[!htbp]
\caption{{Ablation study on AudioSetZSL dataset using different meta-learning methods for $m$ ways $\in$ \{5, 10\} with k-shot $\in$ \{1, 5\}. $\mathcal{X_V}$ and $G_1$($\mathcal{X_A}$) represent the original visual and CMM-GAN hallucinated visual modalities, respectively.}}
\begin{center}
\scalebox{0.63}{
\begin{tabular}{l|c|c c|c c}
\hline
\textbf{Model}&\textbf{m-way}&\multicolumn{2}{c}{Train on $\mathcal{X_V}$ : Test on ${G_1}$$(\mathcal{X_A})$} 
&\multicolumn{2}{|c}{Train on $G_1$$(\mathcal{X_A})$ : Test on $G_1$$(\mathcal{X_A})$}\\
\cline{3-6} 
& &\textbf{1-shot}&\textbf{5-shot}&\textbf{1-shot}&\textbf{5-shot}\\
\hline
\hline

Prototypical \cite{snell2017prototypical} &5-way    &39.54$\pm$0.13 &55.16$\pm$0.44     &38.63$\pm$0.25 &51.09$\pm$0.41\\

Relation Net \cite{sung2018learning} &5-way   &39.72$\pm$0.33 &48.27$\pm$0.34   &46.75$\pm$0.35&47.81$\pm$0.32\\

MAML \cite{DBLP:conf/nips/FinnXL18} &5-way    &43.63$\pm$0.43 &47.32$\pm$0.23     &42.94$\pm$0.32 &45.33$\pm$0.61\\
Matching-Net \cite{matching-net} &5-way    &39.89$\pm$0.30 &49.52$\pm$0.26     &39.60$\pm$0.40 &46.22$\pm$0.34\\

HAVE-Net&5-way        &\textbf{\color{blue}{47.62$\pm$0.24}} &\textbf{\color{blue}{59.11$\pm$0.20}} &\textbf{\color{blue}{48.73$\pm$0.17}} &\textbf{\color{blue}{59.26$\pm$0.22}}\\
\hline

Prototypical \cite{snell2017prototypical} &10-way        &26.50$\pm$0.38 &38.81$\pm$0.34    &25.37$\pm$0.45 &37.65$\pm$0.31\\

Relation Net \cite{sung2018learning} &10-way     &30.08$\pm$0.23 &31.74$\pm$0.44     &28.92$\pm$0.67 &30.59$\pm$0.36 \\

MAML \cite{DBLP:conf/nips/FinnXL18} &10-way    &32.37$\pm$0.27 &40.18$\pm$0.40     &31.00$\pm$0.66 &38.46$\pm$0.34\\
Matching-Net \cite{matching-net} &10-way    &40.36$\pm$0.29 &35.50$\pm$0.33     &28.65$\pm$0.42 &37.11$\pm$0.36\\

HAVE-Net&10-way        &\textbf{\color{blue}{36.91$\pm$0.29}} &\textbf{\color{blue}{43.47$\pm$0.21}}&\textbf{\color{blue}{35.08$\pm$0.23}} &\textbf{\color{blue}{42.68$\pm$0.15}} \\

\hline
\multicolumn{4}{l}{We highlight the best results in \textbf{\color{blue}{blue}}.} \\

\end{tabular}}
\label{abl_Audio}
\end{center}
\end{table}
\noindent{\textbf{Ablation on Effectiveness of CMM-GAN:}} We conduct the evaluations for two different scenarios to further examine the effectiveness of CMM-GAN during inference time; i) The few-shot methods are trained on the original visual modality $\mathcal{X_V}$ and then assessed against the CMM-GAN produced visual modality $G_1$$(\mathcal{X_A})$ and ii) The training and testing of the few-shot approaches with the generated visual modality $G_1$$(\mathcal{X_A})$. Table \ref{abl_adva} and \ref{abl_Audio} illustrate the ablation results for the above-mentioned two circumstances on ADVANCE and AudioSetZSL datasets. We found that HAVE-Net outperforms the above stated scenarios minimum by $4\%$ and $2\%$ on the respective 1-shot and 5-shot settings over both AudioSetZSL and ADVANCE datasets. We show the additional experiments related to model ablation in supplementary paper.

%
%
%
\bibliographystyle{splncs04}
\bibliography{egbib}
%



